# A Comprehensive Evaluation of Machine Learning Techniques for Cancer Class Prediction Based on Microarray Data

Khalid Raza*, Atif N Hasan

*Department of Computer Science, Jamia Millia Islamia (Central University), New Delhi-110025,India*
*kraza@jmi.ac.in

*Abstract*— **Prostate cancer is among the most common cancer in males and its heterogeneity is well known. Its early detection helps making therapeutic decision. There is no standard technique or procedure yet which is full-proof in predicting cancer class. The genomic level changes can be detected in gene expression data and those changes may serve as standard model for any random cancer data for class prediction. Various techniques were implied on prostate cancer data set in order to accurately predict cancer class including machine learning techniques. Huge number of attributes and few number of sample in microarray data leads to poor machine learning, therefore the most challenging part is attribute reduction or non significant gene reduction. In this work we have compared several machine learning techniques for their accuracy in predicting the cancer class i.e., Tumor or Normal. Machine learning is effectively good when the number of attributes (genes) are larger than the number of samples which is rarely possible with gene expression data. Attribute reduction or gene filtering is absolutely required in order to make the data more meaningful as most of the genes do not participate in tumor development and are irrelevant for cancer prediction. Here we have applied combination of statistical techniques such as inter-quartile range and t-test, which has been effective in filtering significant genes and minimizing noise from data. Further we have done a comprehensive evaluation of ten state-of-the-art machine learning techniques for their accuracy in class prediction of prostate cancer. The prostate cancer training data set consists of 12600 genes and 102 samples (instances) and test data set has the same number of genes and 34 different samples and has an overall ten-fold difference from the training data set. After applying inter-quartile range followed by a t-test statistics for attribute reduction we got 856 most significant genes. All the used machine learning techniques were trained with these significant genes. Out of these techniques, Bayes Network out performed with an accuracy of 94.11% followed by Navie Bayes with an accuracy of 91.17%. To cross validate our results, we modified our training dataset in six different way and found that average sensitivity, specificity, precision and accuracy of Bayes Network is highest among all other machine learning techniques used. We also compared our results with the others work on the same kind of dataset and found that our results are better than others.**

## I. INTRODUCTION

TUMOR STATE of prostate cancer is difficult to detect, as prostate cancer is heterogenic in nature [1]. The conventional diagnostic techniques are not always effective as they rely on the physical and morphological appearance of the tumor. Early stage prediction and diagnosis is difficult with those conventional techniques. Moreover, these techniques are also costly, time consuming, requires large laboratory setup and highly skilled persons. It is well known that cancers are involved in genome level changes [2]. Thus, it implies that for a specific type of cancer there could be pattern of genomic change. If those patterns are known, then it can serve as a model for the detection of that cancer [3] and will help in making better therapeutic decisions.

Due to recent advancements in high-throughput techniques for measuring gene expression, it is now possible to monitor the expression levels of tens of thousands of gene at a time. Several researchers have done significant researches, using microarray gene expression data to classify cancers [4] but still predicting cancer class with high accuracy remains a challenge. Here we have done a comparative evaluation of several machine learning techniques for their accuracy in predicting the cancer sample class i.e., Tumor or Normal.

The main difficulty with any machine learning technique is to get trained with large number of genes and comparatively very few samples [5]. This is known as "curse of dimensionality" problem. Machine learning is effectively good when the samples are more and attributes are less but this is rarely possible with gene expression data. Thousands of genes in gene expression data makes the data huge and tough for any machine learning technique to get trained on it. Attribute reduction or gene filtering is absolutely required in order to make the data more meaningful [6] as most of the genes do not participate in tumor development which means that they are irrelevant for cancer prediction. Many researchers have used various techniques for attribute reduction or gene filtering. Here we have applied combination of statistical techniques such as Quartile Range and t-test, which has been effective in filtering significant genes and minimizing noise and irrelevant attributes from data.

We have used ten different machine learning techniques such as *Bayes Network (BN)*, *Naive Bayes (NB)*, *Logistic Model Tree (LMT)*, *C4.5*, *Decision Table (DT)*, *Sequential Minimal Optimization Support Vector Machine* (*SMO-SVM*), *Logit Boost (LB)*, *Random Forest (RF)*, *Neural Network (NN)* and *Genetic Algorithm* (GA).

## II. PREVIOUS RELATED WORKS

Several machine learning techniques [7] such as, Support Vector Machines (SVM) [8], k-Nearest Neighbours (kNN) [9], Artificial Neural Networks [10], Genetic Programming [11], Genetic Algorithms [12], Bayesian Network [13], Naive Bayes [14], Decision Trees [15], Rough Sets [16], Emerging Patterns [17], Self Organising Maps [18], have been used for feature selection, attribute reduction, class prediction and classification using gene expression data. SVM has been used for knowledge based gene expression analysis, recognition of functional classes of genes [15], gene selection [19]. A statistical based method has been used in [43] for identification and filtration of most significant genes that can act as a best drug target. kNN was successfully used for making a model which was capable of predicting the identity of unknown cancer samples [3]. The problem of gene reduction from huge microarray data set was solved by neural network [20], moreover, it was able to identify the genes responsible for particular type of cancer occurrence. Genetic Algorithm was used for building selectors where the state of allele denotes whether it (gene) is selected or not [21]. Genetic Programming has been shown to work excellent in case for recognition of structures for large data sets [22]. It was also applied to microarray data to generate programs that predict the malignant states of cancerous tissue, as well as classify different types of tissues [23]. Bayesian Network were well applied for identification of gene regulatory network from time course microarray data [24]. Self Organising Maps shows good result for gene clustering [25]. Naive Bayes has been used by several researchers for gene selection and classification [26] [27]. Emerging Pattern is markedly good for microarray data analysis. Moreover it has an advantage of designing interaction among genes which enhances classification accuracy [28]. In our recent work, we constructed a prostate cancer-specific gene regulatory network from gene expression profiles and identified some highly connected hub genes using Pearson's correlation coefficient [34]. We also applied information theoretic approach for reconstruction and analysis of gene regulatory network of colon cancer in one of our recent work [42]. But still which method one should apply for the classification and prediction of a particular cancer with high accuracy remains a challenge.

## III. MATERIALS AND METHODS

The prostate cancer data set was taken from Kent Ridge Biomedical Data Repository. The Training data set has 12600 genes (attributes) and 102 samples (instances), while the test data set has the same number of gens and 34 different samples and has an overall ten-fold difference from the training data set. The methodologies used in this study are discussed as follows.

*Data Normalization and Attribute Reduction:* As the available dataset has large number of genes compared to samples, before training machine learning we have done data normalization using inter-quartile range and attribute reduction using t-test. Normalization is a data preprocessing technique used to rescale attribute values to fit in a specific range. Normalizing data is important when dealing with attributes of different units and scales. The inter-quartile range (IQR) is a measure of dispersion which is defined as the difference between the upper and lower quartiles, i.e., IQR = $Q3 - Q1$, where the first quartile $Q1$ represents a quarter and the third quartile $Q3$ three quarters of the list of all the data. The IQR is essentially the range of the middle 50% of the data and hence it is not affected by outliers or extreme values.

After normalizing data we used a two tailed t-test for extracting differentially expressed genes among two types of sample, i.e., Normal and Tumor, at a significant level $\alpha=0.001$. General formula of t-Test of unequal sample size is,

$$t = \frac{\overline{x}_1 - \overline{x}_2}{S_{x_1 x_2} \sqrt{\frac{1}{n_1} + \frac{1}{n_2}}} \quad (1)$$

where,

$$S_{x_1 x_2} = \sqrt{\frac{(n_1 - 1)S_{x_1}^2 + (n_2 - 1)S_{x_2}^2}{n_1 + n_2 - 2}} \quad (2)$$

$x_1$ and $x_2$ are two unequal samples and $\overline{x}_1 - \overline{x}_2$ is the difference between the sample mean. $S_{x_1 x_2}$ is the standard error of the difference, $n_1$ and $n_2$ are the size of the samples. In this case we have $n_1$ and $n_2$ as 52 and 50 for Tumor and Normal samples, respectively. After applying t-test, out of 12600 genes, 856 genes were extracted out. These genes are the differentially expressed genes. From the test data set, we extracted out the same genes as in remained in the training data set after normalization and attribute reduction.

*1. Naive Bayes:* It is a simple probability based techniques mainly based on Bayes theorem with high independence assumption [30]. The presence or absence of any attribute is not dependent on other. It requires small amount of training data in order to estimate the parameters required for classification [31][32]. The probability of posterior ($p$) which depends over a class variable $C$ conditional on variable features $F_1,…F_n$, where $n$ is the number of features, is given by:

$$p(C | F_1,...,F_n) = \frac{p(C)p(F_1,...,F_n | C)}{p(F_1,...,F_n)} \quad (3)$$

*2. Bayes Net:* Bayes Net was developed for improving the performance of Naive Bayes. These are the directed acyclic graphs (DAG) which allow an effective representation of joint probability distribution for a set of random variables [30]. A finite set $\mathbf{U} = \{X_1,...,X_n\}$, of discrete random variables $X_i$, then the joint probability distribution $P_B$ over the variables of set $\mathbf{U}$ is given by:

$$P_B(X_1,...,X_n) = \prod_{i=1}^{n} P_B(X_i \mid \prod X_i) = \prod_{i=1}^{n} \theta_{X_i \mid \prod X_i} \qquad (4)$$

A training set D = {t_1,...,t_n} of instances of T, finds a network B which best matches the training set D.

*3. LogitBoost:* Boosting was well describe by "Freund and Schapire" that it is a classification which works by sequential implementation of a classification algorithm to reweighted training data and then taking the sequence classifiers produced by the weighted majority vote [33]. For two class problem boosting can be taken as an approximation to additive modeling on the logistic scale based on Bernoulli likelihood as a criterion. When cost function of logistic regression is applied on generalized model of AdaBoost, LogitBoost is derived.

*4. C4.5:* It is an extension of ID3 algorithm proposed by Ross Quinlan [35] which is used to generate a decision tree for classification. C4.5 is also known as statistical classifier. The C4.5 constructs decision tree from a set of training dataset using information entropy concepts. If we have training dataset $S=\{s_1, s_2, s_3,...\}$ which are already classified samples, each sample $s_i$ consists of a p-dimensional vector $(x_{1i}, x_{2i}, ..., x_{pi})$, where $x_j$ stands for attributes of the sample and the class in which sample $s_i$ falls. At every node of the tree, C4.5 chooses the attribute that most effectively divides its set of samples into subsets. The division condition is based on normalized information gain, i.e., difference in entropy. The attribute having highest normalized information gain is selected to take decision. The C4.5 algorithm then recursively works on the smaller sub-lists. C4.5 avoids over-fitting of data, determines how deeply a decision tree would grow, reduces error pruning, rule post-pruning, handles continuous attributes, choosing a suitable attribute selection measure, handles training data with missing values and improves computational efficiency.

*5. Logistic Model Trees:* This is the combined version of linear logistic regression and tree induction. The former produces low variance high bias and the later produces high variance low bias. These two techniques were combined into learner which depends upon simple regression models if only little and/or noisy data is present. It adds more complex tree structures if enough data is available to such structures. Thus logistic model trees are the decision trees having linear regression model at leaves [36].

*6. Random Forest*: It is type of ensemble learning classification method. During training it constructs many decision trees. Mode class is extracted which is the mode of the classes output by individual trees [37]. Random vectors are generated which leads to the growth of individual trees in the ensemble. As the definition given by L. Breiman, 2001, *"Random forest is a classifier consisting of a collection of tree-structured classifiers {h(x, θk), k = 1,...} where the {θk} are independent identically distributed random vectors and each tree casts a unit vote for the most popular class at input x".*

*7. Decision Table:* It is an easy way to model complicated logics. These are flowcharts based on if, then, else, switch cases statements and associate conditions with actions to perform. Each decision is related to a variable, relation, condition alternatives dependencies. Operations to be performed are actions which correspond to specific entry. Each entry specify whether or in what order the action is to be performed for the given set of condition alternatives the entry corresponds to [38].

*8. SMO-SVM*: Sequential minimal optimization (SMO) is an efficient algorithm which solves the optimization problem of Support Vector Machine (SVM) which arises during training. It breaks the optimization problem into several sub-problems, which are then solved analytically. The larger multiplier $α_1$ of the problem has linear equality constraint and therefore the smallest possible problem has two such multipliers. Then for any two multiplier ($α_1$ and $α_2$), the constraints are reduced to $0 \leq α_1, α_2 \leq C$, $y_1 α_1 + y_2 α_2$ and solved analytically [39].

*9. Neural Network:* It is similar to the biological neurons. The artificial neurons or nodes makes it artificial/simulated neural network. The group of interconnected nodes processes information using computational and mathematical model. Based on the information (external or internal) flowing through the network the adaptive system of artificial neural network changes its structure. Neural network are nonlinear statistical data modeling tools [40].

*10. Genetic Algorithm:* In the field of artificial intelligence, genetic algorithm is heuristic search algorithm that mimics the natural process of evolution. It works well for optimization and search problem using techniques such as inheritance, mutation, selection, crossover, which are inspired by natural evolution [41]. Genetic representation and fitness function is set and the process proceeds for initialization, where for a given population several solutions are randomly generated. For each successive generation, a portion from existing population is selected to breed a new generation. This constitutes the selection process, which is based on fitness selection process. With the help of operators like crossover and mutation new generation is generated. This process repeated until a termination condition is achieved [41].

## IV. RESULTS AND DISCUSSIONS

The experiment was carried on prostate cancer data set taken from Kent Ridge Biomedical Data Repository (http://datam.i2r.a-star.edu.sg/datasets/krbd/). Data was collected from the pioneer publication of Dinesh Singh. et al. [3]. In his experiment 235 radical prostatectomy specimens were analyzed from different patients. From that, 65 samples specimen were identified for having tumor on opposing side of the tissue. Gene expression profile was successfully carried for 52 prostate tumor and 50 non-tumor prostate samples containing expression profile of 12600 genes. This makes the training data set for our experiment. Whereas the test data set was having nearly ten-fold difference in over all microarray intensity form the training data set and was taken from independent experiment. It has 25 tumor samples and 9 non tumor/normal samples. We have used independent data set for measuring the accuracy of different techniques because training on one data and testing on other independent data is more reliable. Our work is different from others because they have used the same data for training and a percent split of it for testing.

The non reduced or non filtered data was full of noise and irrelevant data. The maximum and minimum values within the data prior to attribute reduction were 17530 and -1807 respectively, therefore the range was 19337. After attribute reduction maximum and minimum values were 18.06729562 and -13.4414901 respectively, and therefore the range reduced to 31.508785. Thus the range after attribute reduction is far less prior to attribute reduction and shows that reduced data is less scattered. This satisfies the basic requirement for machine learning techniques. Table 1 shows a part of training set before attribute reduction, whereas Table 2 shows a part training data set after attribute reduction.

**Table 1**
TRAINING DATA SET BEFORE ATTRIBUTE REDUCTION

| Gene 1 | Gene 2 | Gene 3 | Gene 4 | Gene 5 | Gene 6 |
| --- | --- | --- | --- | --- | --- |
| -9 | 1 | 1 | 15 | -2 | -3 |
| -2 | 1 | 1 | 4 | -2 | -5 |
| -6 | 17 | 6 | 29 | 4 | -11 |
| 0 | 9 | 4 | 19 | -10 | -18 |
| -1 | 0 | 1 | 5 | 0 | -4 |
| 0 | 17 | 1 | 20 | -20 | -18 |
| -5 | 5 | -1 | 9 | -10 | -17 |
| -3 | 1 | 1 | 5 | -2 | -6 |
| -8 | -2 | -1 | -32 | -20 | -41 |
| -12 | 11 | -3 | 21 | -10 | -9 |

The table shows the first six genes in columns and their corresponding gene expression values in first ten samples. The values of genes are largely scattered throughout the matrix. This expression matrix is raw, non reduced and non filtered.

**Table 2**
TRAINING DATA SET AFTER ATTRIBUTE REDUCTION

| Gene 1 | Gene 2 | Gene 3 | Gene 4 | Gene 5 | Gene 6 |
| --- | --- | --- | --- | --- | --- |
| -0.14 | 0.30 | 0.89 | -1.11 | 0.19 | 0.56 |
| -0.14 | 0.39 | 0.66 | -1.79 | -0.41 | 0.31 |
| 0.05 | -0.36 | -1.13 | -0.59 | -0.56 | -0.99 |
| -0.43 | -1.04 | -0.65 | -0.20 | 1.54 | -0.92 |
| -0.34 | 0.04 | -0.03 | 0.07 | -0.41 | -0.03 |
| 0.53 | 0.62 | -0.99 | -0.38 | -0.56 | -0.60 |
| 0.34 | -0.62 | -0.08 | -0.64 | 0.12 | -0.77 |
| -0.92 | 0.87 | -0.37 | -1.20 | -0.71 | 0.32 |
| 1.98 | -1.51 | -1.37 | 2.25 | 1.69 | 0.12 |
| 0.24 | -0.97 | -1.05 | -0.51 | 0.19 | 0.06 |

The table shows the first six genes in columns after attribute reduction and their corresponding gene expression values in first ten samples. Difference between the highest and the lowest values of genes are far less as compared to non reduced values.

Following are the description of the results of various techniques used for training and testing. The training data set has 102 samples where as the test data set has 34 samples. Total number of Tumor sample and Normal sample in test data set are 25 and 9 respectively (Here after correctly classified samples and incorrectly classified samples will be denoted as CCS and ICS, respectively.) Table 3 shows a brief performance comparison of different techniques used with different statistical measures. It is clearly depicted that Bayes Network outperforms as compared to other techniques. Figure 1 shows the comparison of different techniques used for their accuracy in classifying samples correctly and incorrectly, whereas Figure 2 shows accuracy level of different techniques used.

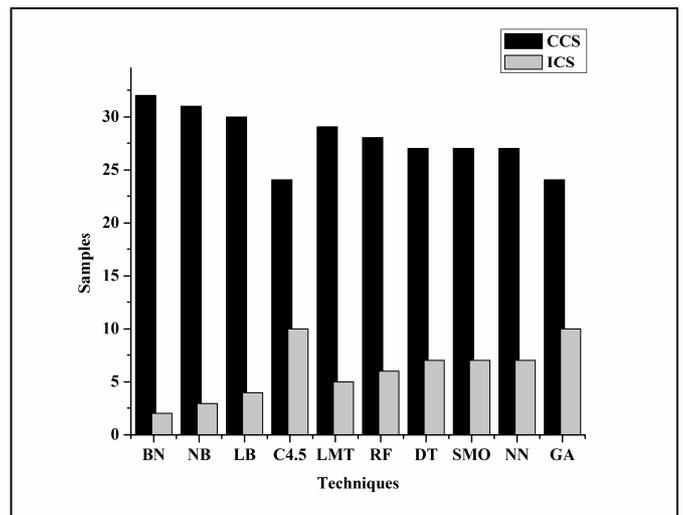

**Fig. 1. Correctly Classified samples (CCS) versus Incorrectly Classified Samples (ICS).** While testing, total number of samples were 34. Bayes net was found to be the best technique for classifying cancer class. Out of 34 samples Bayes net classified 32 samples correctly.

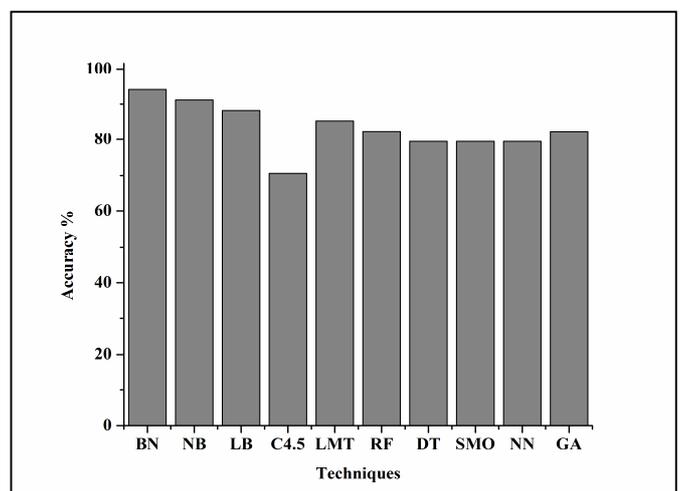

**Fig. 2. Accuracy level of various techniques on test data set. BN** (Bayes Net ), **NB** (Naive Bayes), **LB** (Logit Boost), **C4.5**, **LMT** (Logistic Model Tree), **RF** (Random Forest), **DT** (Decision Table), **SMO-SVM** (Sequential Minimal Optimization- Support Vector Machine), **NN** (Neural Network), **GA** (Genetic Algorithm). Bayes Net outperforms the other techniques with an accuracy level of 94.11% followed by Naive Bayes with an accuracy level of 91.17%. Instance Based and Logistic Model Tree have same accuracy level of 85.29%, Random Forest and Genetic Algorithm have the same accuracy level of 82.35%, Decision Tree and SMO-SVM have same accuracy level of 79.41%. C4.5 has the lowest accuracy level of 70.58%.

## OUTCOMES OF THE TECHNIQUES USED

*1. Bayes Net*: While training CCS was 93 and ICS was 9. Percent accuracy was 91.17%. During testing CCS was 32 and ICS was 2. Out of 25 Tumor samples 23 were detected as Tumor and out of 9 Normal samples, all 9 were detected as Normal. Percent accuracy was 94.11%.

*2. Naive Bayes*: While training CCS was 89 and ICS was 13. Percent accuracy was 87.25%. During testing CCS was 31 and ICS was 3. Out of 25 Tumor samples 22 were detected as Tumor and out of 9 Normal samples, all 9 were detected as Normal. Percent accuracy was 91.17%.

*3. Logit Boost*: While training CCS was 102 and ICS was 0 Percent accuracy was 100.0%. During testing CCS was 30 and ICS was 4. Out of 25 Tumor samples 21 were detected as Tumor and out of 9 Normal samples 9 were detected as Normal. Percent accuracy was 88.23%.

*4. C4.5:* While training CCS was 101 and ICS was 1. Percent accuracy was 99.01%. During testing CCS was 24 and ICS was 10. Out of 25 Tumor samples 19 were detected as Tumor and out of 9 Normal samples 5 were detected as Normal. Percent accuracy was 70.58%.

*5. Logistic Model Trees:* While training CCS was 102 and ICS was 0. Percent accuracy was 100.0%. During testing CCS was 29 and ICS 5 was. Out of 25 Tumor samples 20 were detected as Tumor and out of 9 Normal samples 9 were detected as Normal. Percent accuracy was 85.29 %.

*6. Random Forest*: While training CCS was 101 and ICS was 1.0. Percent accuracy was 99.01%. During testing CCS was 28 and ICS was 6. Out of 25 Tumor samples 19 were detected as Tumor and out of 9 Normal samples 9 were detected as Normal. Percent accuracy was 82.35%.

*7. Decision Table:* While training CCS was 99 and ICS was 3. Percent accuracy was 97.05. During testing CCS was 27 and ICS was 7. Out of 25 Tumor samples 19 were detected as Tumor and out of 9 Normal samples 8 were detected as Normal. Percent accuracy was 79.41%.

*8. SMO-SVM*: While training CCS was 102 and ICS was 0 Percent accuracy was 100.0%. During testing CCS was 27 and ICS was 7. Out of 25 Tumor samples 18 were detected as Tumor and out of 9 Normal samples 9 were detected as Normal. Percent accuracy was 79.41%.

*9. Neural Network:* While training the minimum and final Root Mean Squared Error (RMSE) was 0.1277. while testing RMSE was 0.2536. During testing CCS was 27 and ICS was 7. Out of 25 Tumor samples 18 were detected as Tumor and out of 9 Normal samples 9 were detected as Normal. Percent accuracy was 79.41%.

*10. Genetic Algorithm*: While training the minimum Mean Squared Error (MSE) for Best Fitness and Average Fitness was 2.91173E-05 and 0.006661297 respectively. The final MSE for Best Fitness and Average Fitness was 2.91173E-05 and 0.038575474 respectively. During testing the Average root mean squared error was 0.339.3 where as CCS was 28 and ICS 6 was. Out of 25 Tumor samples 19 were detected as Tumor and out of 9 Normal samples 9 were detected as Normal. Percent accuracy was 82.35. Figure 3 shows the Average fitness versus generation graph while testing. The best fitness was for the 7$^{th}$ and 8$^{th}$ generation with mean square error (MSE) less than 0.05.

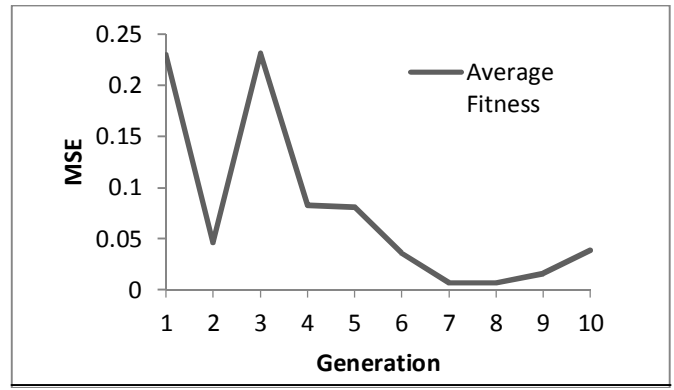

**Fig. 3**. Average fitness versus generation graph. Among the ten generation the best fitness was for the 4$^{th}$ generation with mean square error (MSE) less than 0.05. The average MSE is 0.164830.

Table 3
PERFORMANCE COMPAIRISION OF TECHNIQUE

| Techniques | CCS | ICS | RMSE | TPR (Tumor) | TPR (Normal) | FPR (Tumor) | FPR (Normal) |
|---|---|---|---|---|---|---|---|
| Bayes Net | 32 (94.11%) | 2 (8.82%) | 0.2189 | 0.92 | 1 | 0 | 0.8 |
| Naive Bayes | 31 (91.17%) | 3 (8.82%) | 0.297 | 0.88 | 1 | 0 | 0.12 |
| Logit Boost | 30 (88.23%) | 4 (11.76%) | 0.3736 | 0.84 | 1 | 0 | 0.16 |
| C4.5 | 24 (70.58%) | 10 (29.41%) | 0.4918 | 0.76 | 0.55 | 0.44 | 0.24 |
| Logistic Model Tree | 29 (85.29%) | 5 (14.70%) | 0.3429 | 0.8 | 1 | 0 | 0.2 |
| Random Forest | 28 (82.35%) | 6 (17.64%) | 0.3523 | 0.76 | 1 | 0 | 0.24 |
| Decision Table | 27 (79.41%) | 7 (20.58%) | 0.3523 | 0.76 | 0.889 | 0.111 | 0.24 |
| SMO-SVM | 27 (79.41%) | 7 (20.58%) | 0.4537 | 0.72 | 1 | 0 | 0.28 |
| Neural Network | 27 (79.41%) | 7 (20.58%) | 0.253693 | 0.88 | 1 | 0 | 0.12 |
| Genetic Algorithm | 24 (76.44%) | 10 (29.41%) | 0.3747 | 0.94 | 0.47 | 0.52 | 0.058 |

**CCS** (correctly classified samples), **ICS** (incorrectly classified samples), **RMSE** (root mean squared error), **TPR** (true positive rate), **FPR** (false positive rate). Bayes Net gives the most accurate prediction of prostate cancer class with an accuracy of 94.11%. Total number of sample was 34 out of which it predicted 32 samples correctly.

## CROSS VALIDATION BY MODIFYING TEST DATA SETS

As the test data set was taken from independent experiments and was having a overall tenfold difference from the training data set, we have modified the test data set in six different ways to cross validate our results. The test data set was divided by 2 (Div 2), 10 (Div 10), 20 (Div 20) and multiplied by 2 (Mul 2), 10 (Mul 10), 20 (Mul 20), making six different data sets. Table 4 shows the accuracy level of different techniques for all the six modified test data as well as the average accuracy level. It is clear from the table that the average accuracy level of Bayes Net is more than other techniques.

**Sensitivity and specificity analysis**

Sensitivity is one of the statistical methods for measuring the performance of binary classification [29]. It measures the True Positive rate. True Positives (TP) are the positives which are correctly identified as positive. A high sensitivity corresponds to higher accuracy. Whereas, specificity describes the ability (of any technique used) to identify negatives as negative or true negatives (TN) as true negative. Thus, a high specificity indicates that any technique used has a high ability to identify true negatives. General formula for sensitivity and specificity are,

$$Sensitivity = \frac{TP}{TP + FN} \quad (5)$$

$$Specificity = \frac{TN}{TN + FP} \quad (6)$$

Table 4 shows sensitivity and specificity of all the techniques used for all the six modified data. Sensitivity and specificity of all the techniques was calculated for all the six modified data. Then the actual (mean) sensitivity and specificity of a technique was calculated by averaging the sensitivity and specificity obtained for individual modified data sets. Table 5 shows the actual (mean) sensitivity and specificity of all the techniques. The outcome shows Bayes Net has the sensitivity and specificity of 0.89 and 1 respectively, which is higher than the others.

**Precision and accuracy analysis**

Precision is also known as reproducibility or repeatability. It shows how a measurement under repeating condition remains unchanged. Precision is the degree of measurement of true positive against true positive and false positive. Whereas accuracy is the degree of closeness of obtained value to the actual value. A high accuracy and high precision signifies that testing process is working well with a valid theory. General formula for precision and accuracy is,

$$Precision = \frac{TP}{TP + FP} \quad (7)$$

$$Accuracy = \frac{TP + TN}{TP + TN + FP + FN} \quad (8)$$

Precision and accuracy of all the techniques were calculated for all the six modified data set. Then the actual precision and accuracy of a technique was calculated by taking the mean of precisions and accuracies obtained for all the modified data sets. Table 6 shows the actual (mean) precision and accuracy of all the techniques for modified datasets. It shows clearly that Bayes Net has the highest precision and accuracy of 1 and 0.92, respectively.

**Table 4**
SENSITIVITY AND SPECIFICITY

| Techniques | Sn/Sp | Div 2 | Mul 2 | Div 10 | Mul 10 | Div 20 | Mul 20 |
|---|---|---|---|---|---|---|---|
| BN | Sn | 0.96 | 0.92 | 0.88 | 0.8 | 1 | 0.8 |
|  | Sp | 1 | 1 | 1 | 1 | 1 | 1 |
| NB | Sn | 1 | 0.52 | 1 | 0.2 | 1 | 0.12 |
|  | Sp | 0.88 | 1 | 0 | 1 | 0 | 0.88 |
| LB | Sn | 0.92 | 0.76 | 0.92 | 0.68 | 1 | 0.68 |
|  | Sp | 1 | 1 | 0.44 | 1 | 0.11 | 1 |
| C4.5 | Sn | 0.64 | 0.76 | 0.68 | 0.84 | 0.68 | 0.84 |
|  | Sp | 1 | 0.55 | 0.66 | 0.55 | 0.66 | 1 |
| LMT | Sn | 0.84 | 0.76 | 1 | 0.64 | 1 | 0.64 |
|  | Sp | 1 | 1 | 0 | 1 | 0 | 1 |
| RF | Sn | 0.92 | 0.72 | 0.92 | 0.48 | 1 | 0.44 |
|  | Sp | 1 | 1 | 0.66 | 0.77 | 0.44 | 0.77 |
| DT | Sn | 0.76 | 0.72 | 0.68 | 0.72 | 0.68 | 0.72 |
|  | Sp | 0.88 | 0.88 | 0.66 | 0.88 | 0.55 | 0.88 |
| SMO | Sn | 0.8 | 0.64 | 1 | 0.64 | 1 | 0.64 |
|  | Sp | 1 | 1 | 0 | 1 | 0 | 1 |
| NN | Sn | 0.8 | 0.76 | 1 | 0.72 | 0 | 0.84 |
|  | Sp | 1 | 1 | 0 | 1 | 1 | 1 |
| GA | Sn | 0.68 | 0.72 | 0.68 | 0.76 | 1 | 0.72 |
|  | Sp | 0.88 | 1 | 1 | 0 | 0.88 | 1 |

Sn = Sensitivity and Sp = Specificity. Table shows sensitivity and specificity of different techniques for all the six modified data sets.

**Table 5**
MEAN SENSITIVITY AND SPECIFICITY COMPARISION

| Techniques | BN | NB | LB | C4.5 | LMT | RF | DT | SMO | NN | GA |
|---|---|---|---|---|---|---|---|---|---|---|
| Sensitivity | 0.89 | 0.64 | 0.82 | 0.74 | 0.81 | 0.74 | 0.71 | 0.78 | 0.68 | 0.76 |
| Specificity | 1 | 0.62 | 0.75 | 0.73 | 0.66 | 0.77 | 0.78 | 0.66 | 0.83 | 0.79 |

Table shows the mean sensitivity and means specificity of all the techniques. The means were calculated by averaging the sensitivity and specificity of all the techniques for all the six modified data sets. Bayes Net has the highest sensitivity and specificity of 0.89 and 1, respectively.

**Table 6**
PRECISION AND ACCURACY COMPARISION

| Techniques | BN | NB | LB | C4.5 | LMT | RF | DT | SMO | NN | GA |
|---|---|---|---|---|---|---|---|---|---|---|
| Precision | 1 | 0.86 | 0.92 | 0.77 | 0.71 | 0.71 | 0.71 | 0.76 | 0.7 | 0.7 |
| Accuracy | 0.92 | 0.63 | 0.8 | 0.74 | 0.77 | 0.75 | 0.73 | 0.75 | 0.7 | 0.7 |

Table shows the comparison of precision and accuracy. Bayes Net has the highest precision and accuracy.

In the following section Table 7 shows a comparison of classification accuracy of our work with the works of other researchers on the same kind of dataset (prostate cancer). Our Bayesian network and Naive Bayes based techniques shows the highest accuracy over the others, i.e., an accuracy of 94.11 and 92.17, respectively. The kNN based method of D. Singh et al [3] shows the next highest accuracy between 86-92.

**Table 7**
COMPARISION OF ACCURACY WITH THE WORK OF OTHERS

| Author(s) | Techniques | Accuracy (%) |
|---|---|---|
| **Our method** | **Bayesian network** | **94.11** |
| | **Naive Bayes** | **92.17** |
| B. Zupan et al [44] | Naive Bayes | 70.8 - 78.4 |
| | Decision Tree | 68.8 -77 |
| | Cox | 69.7 -79 |
| M. Wagner et al [45] | kNN | 87.4 - 89.9 |
| | Fisher Linear | 87.9 - 89.1 |
| | Linear SVM | 89.5 - 91.9 |
| Tan & Gilbert [46] | Single C4.5 | 52.38 |
| | Bagging C4.5 | 85.71 |
| | AdaBoost | 76.19 |
| D. Singh et al [3] | kNN | 86 - 92 |

Table shows the comparison of classification accuracy with the works of others on prostate cancer dataset. Our Bayesian network and Naive Bayes based classification method outperforms over the others.

## V. CONCLUSIONS & FUTURE CHALLENGES

In this paper we have comparatively evaluated various machine learning techniques for their accuracy in class prediction of prostate cancer data set. As per our evaluation, Bayes Net gave the best accuracy for prostate cancer class prediction with an accuracy of 94.11% which is higher than any previously published work on the same data set. Bayes Net is followed by Navie Bayes with an accuracy of 91.17%. We tested our data set on different techniques and selected those techniques which gave best results. Our aim was to identify the best technique in terms of accuracy which can classify prostate cancer date set and to reveal a good procedure for meaningful attribute reduction, which we have acquired by using a combination of t-test and inter-quartile range. Similar process can be applied and checked for their accuracy in classification of other types of cancers. One of the biggest challenges is to develop a single classifier which is best suitable for classifying all types of cancer gene expression data into meaningful number of classes. Nature inspired optimization techniques such as Ant Colony Optimization (ACO), Artificial Be Colony optimization (ABC), Particle Swarm Optimization (PSO) are successfully being used in many challenging problems. In the future work, we are willing to hybridized these nature inspired optimization techniques with different classifiers for better classification accuracy.

## ACKNOWLEDGEMENTS

The authors would like to thank all the scientist behind Kent Ridge Biomedical Data Set Repository for making the datasets publicly available. The author K. Raza acknowledges the funding from University Grants Commission, Govt. of India through research grant 42-1019/2013(SR).

## REFERENCES

[1] M. Aihara, T. M. Wheeler, M. Ohori, and P. T. Scardino, "Heterogeneity of prostate cancer inradical prostatectomy specimens," *Urology*, vol. 43, no. 1, pp. 60–66, Jan. 1994.
[2] M. R. Stratton, P. J. Campbell, and P. A. Futreal, "The cancer genome," *Nature*, vol. 458, no. 7239, pp. 719–724, Apr. 2009.
[3] D. Singh, P. G. Febbo, K. Ross, D. G. Jackson, J. Manola, C. Ladd, P. Tamayo, A. a Renshaw, A. V D'Amico, J. P. Richie, E. S. Lander, M. Loda, P. W. Kantoff, T. R. Golub, and W. R. Sellers, "Gene expression correlates of clinical prostate cancer behavior.," *Cancer cell*, vol. 1, no. 2, pp. 203–9, Mar. 2002.
[4] T. R. Golub, D. K. Slonim, P. Tamayo, C. Huard, M. Gaasenbeek, J. P. Mesirov, H. Coller, M. L. Loh, J. R. Downing, M. a Caligiuri, C. D. Bloomfield, and E. S. Lander, "Molecular classification of cancer: class discovery and class prediction by gene expression monitoring.," *Science (New York, N.Y.)*, vol. 286, no. 5439, pp. 531–7, Oct. 1999.
[5] Y. Saeys, I. Inza, and P. Larrañaga, "A review of feature selection techniques in bioinformatics.," *Bioinformatics (Oxford, England)*, vol. 23, no. 19, pp. 2507–17, Oct. 2007.
[6] J. Quackenbush, "Microarray data normalization and transformation.," *Nature genetics*, vol. 32 Suppl, no. december, pp. 496–501, Dec. 2002.
[7] Y. Lu and J. Han, "Cancer classification using gene expression data," *Information Systems*, vol. 28, no. 4, pp. 243–268, Jun. 2003.
[8] I. Guyon, J. Weston, S. Barnhill, and V. Vapnik, "Gene Selection for Cancer Classification using Support Vector Machines," *Machine Learning*, vol. 46, no. 1–3, pp. 389–422 LA – English, 2002.
[9] S. Varambally, J. Yu, B. Laxman, D. R. Rhodes, R. Mehra, S. A. Tomlins, R. B. Shah, U. Chandran, F. A. Monzon, M. J. Becich, J. T. Wei, K. J. Pienta, D. Ghosh, M. A. Rubin, and A. M. Chinnaiyan, "Integrative genomic and proteomic analysis of prostate cancer reveals signatures of metastatic progression," *Cancer Cell*, vol. 8, no. 5, pp. 393–406, Nov. 2005.
[10] J. Khan, J. S. Wei, M. Ringnér, L. H. Saal, M. Ladanyi, F. Westermann, F. Berthold, M. Schwab, C. R. Antonescu, C. Peterson, and P. S. Meltzer, "Classification and diagnostic prediction of cancers using gene expression profiling and artificial neural networks.," *Nature medicine*, vol. 7, no. 6, pp. 673–9, Jun. 2001.
[11] L. Vanneschi, A. Farinaccio, G. Mauri, M. Antoniotti, P. Provero, and M. Giacobini, "A comparison of machine learning techniques for survival prediction in breast cancer.," *BioData mining*, vol. 4, no. 1, p. 12, Jan. 2011.
[12] T. Jirapech-Umpai and S. Aitken, "Feature selection and classification for microarray data analysis: evolutionary methods for identifying predictive genes.," *BMC bioinformatics*, vol. 6, p. 148, Jan. 2005.
[13] N. Friedman, M. Linial, I. Nachman, and D. Pe'er, "Using Bayesian networks to analyze expression data.," *Journal of computational biology : a journal of computational molecular cell biology*, vol. 7, no. 3–4, pp. 601–20, Jan. 2000.
[14] Y. Wang, I. V Tetko, M. A. Hall, E. Frank, A. Facius, K. F. X. Mayer, and H. W. Mewes, "Gene selection from microarray data for cancer classification—a machine learning approach," *Computational Biology and Chemistry*, vol. 29, no. 1, pp. 37–46, Feb. 2005.
[15] M. P. S. Brown, W. N. Grundy, D. Lin, N. Cristianini, C. W. Sugnet, T. S. Furey, M. Ares, and D. Haussler, "Knowledge-based analysis of microarray gene expression data by using support vector machines," *Proceedings of the National Academy of Sciences*, vol. 97, no. 1, pp. 262–267, 2000.
[16] X. Wang and O. Gotoh, "Microarray-based cancer prediction using soft computing approach.," *Cancer informatics*, vol. 7, pp. 123–39, Jan. 2009.
[17] J. Li and L. Wong, "Identifying good diagnostic gene groups from gene expression profiles using the concept of emerging patterns," *Bioinformatics*, vol. 18, no. 5, pp. 725–734, 2002.
[18] A. L. Hsu, S.-L. Tang, and S. K. Halgamuge, "An unsupervised hierarchical dynamic self-organizing approach to cancer class discovery and marker gene identification in microarray data," *Bioinformatics*, vol. 19, no. 16, pp. 2131–2140, 2003.
[19] Y. Tang, Y.-Q. Zhang, and Z. Huang, "Development of two-stage SVM-RFE gene selection strategy for microarray expression data analysis," *Computational Biology and Bioinformatics, IEEE/ACM Transactions on*, vol. 4, no. 3, pp. 365–381, 2007.
[20] M. O'Neill and L. Song, "Neural network analysis of lymphoma microarray data: prognosis and diagnosis near-perfect," *BMC Bioinformatics*, vol. 4, no. 1, p. 13, 2003.
[21] J. J. Liu, G. Cutler, W. Li, Z. Pan, S. Peng, T. Hoey, L. Chen, and X. B. Ling, "Multiclass cancer classification and biomarker discovery using GA-based algorithms," *Bioinformatics*, vol. 21, no. 11, pp. 2691–2697, 2005.

[22] J. H. Moore, J. S. Parker, and L. W. Hahn, "Symbolic discriminant analysis for mining gene expression patterns," in *Machine Learning: ECML 2001*, Springer, 2001, pp. 372–381.
[23] M. Rosskopf, H. A. Schmidt, U. Feldkamp, and W. Banzhaf, *Genetic Programming based DNA Microarray Analysis for Classification of Cancer*. Memorial University, 2007.
[24] M. Zou and S. D. Conzen, "A new dynamic Bayesian network (DBN) approach for identifying gene regulatory networks from time course microarray data," *Bioinformatics*, vol. 21, no. 1, pp. 71–79, 2005.
[25] A. Sturn, J. Quackenbush, and Z. Trajanoski, "Genesis: cluster analysis of microarray data," *Bioinformatics*, vol. 18, no. 1, pp. 207–208, 2002.
[26] T. Li, C. Zhang, and M. Ogihara, "A comparative study of feature selection and multiclass classification methods for tissue classification based on gene expression," *Bioinformatics*, vol. 20, no. 15, pp. 2429–2437, 2004.
[27] K. Y. Yeung, R. E. Bumgarner, and A. E. Raftery, "Bayesian model averaging: development of an improved multi-class, gene selection and classification tool for microarray data," *Bioinformatics*, vol. 21, no. 10, pp. 2394–2402, 2005.
[28] A.-L. Boulesteix, G. Tutz, and K. Strimmer, "A CART-based approach to discover emerging patterns in microarray data," *Bioinformatics*, vol. 19, no. 18, pp. 2465–2472, 2003.
[29] F. Provost and T. Fawcett, "Analysis and visualization of classifier performance: Comparison under imprecise class and cost distributions," in *Proceedings of the third international conference on knowledge discovery and data mining*, 1997, pp. 43–48.
[30] C. S. Division, M. Park, P. Langley, and P. Smyth, "Bayesian Network Classifiers *," vol. 163, pp. 131–163, 1997.
[31] H. Zhang, "The Optimality of Naive Bayes," 2004.
[32] I. Rish, "An empirical study of the naive Bayes classifier," in *IJCAI 2001 workshop on empirical methods in artificial intelligence*, 2001, vol. 3, no. 22, pp. 41–46.
[33] J. Friedman, T. Hastie, and R. Tibshirani, "Additive logistic regression: a statistical view of boosting (With discussion and a rejoinder by the authors)," *The annals of statistics*, vol. 28, no. 2, pp. 337–407, 2000.
[34] K. Raza and R. Jaiswal. "Reconstruction and Analysis of Cancer-specific Gene Regulatory Networks from Gene Expression Profiles," *International Journal on Bioinformatics & Biosciences*, vol. 3, issue 2, pp. 25-34, June 2013 [Preprint arXiv:1305.5750, 2013].
[35] J.R. Quinlan. C4.5: Programs for Machine Learning. Morgan Kaufmann Publishers, 1993.
[36] N. Landwehr, M. Hall, and E. Frank, "Logistic model trees," in *Machine Learning: ECML 2003*, Springer, 2003, pp. 241–252.
[37] L. Breiman, "Random Forests," *Machine Learning*, vol. 45, no. 1, pp. 5–32 LA – English, 2001.
[38] B. J. Cragun and H. J. Steudel, "A decision-table-based processor for checking completeness and consistency in rule-based expert systems," *International Journal of Man-Machine Studies*, vol. 26, no. 5, pp. 633–648, May 1987.
[39] J. Platt, "Sequential minimal optimization: A fast algorithm for training support vector machines," 1998.
[40] J. J. Hopfield, "Neural networks and physical systems with emergent collective computational abilities," *Proceedings of the national academy of sciences*, vol. 79, no. 8, pp. 2554–2558, 1982.
[41] M. Mitchell, "An introduction to genetic algorithms, 1996," *PHI Pvt. Ltd., New Delhi*, 1996.
[42] K. Raza and R. Parveen, "Reconstruction of gene regulatory network of colon cancer using information theoretic approach", *arXiv*:1307.3712, 2013.
[43] K. Raza and A. Mishra, "A novel anticlustering filtering algorithm for the prediction of genes as a drug target", *American Journal of Biomedical Engineering*, vol. 2, issue 5, pp. 206–211, 2012.
[44] B. Zupan, J. Demsar, M. W. Kattan, J. R. Beck, and I. Bratko, "Machine learning for survival analysis: a case study on recurrence of prostate cancer.," *Artificial intelligence in medicine*, vol. 20, no. 1, pp. 59–75, Aug. 2000.
[45] M. Wagner, D. N. Naik, A. Pothen, S. Kasukurti, R. R. Devineni, B. Adam, O. J. Semmes, and G. L. W. Jr, "Computational protein biomarker prediction : a case study for prostate cancer," vol. 9, pp. 1–9, 2004.
[46] A. C. Tan and D. Gilbert, "data for cancer classification," vol. 2, pp. 1–10, 2003.